\pdfoutput=1

\documentclass[11pt]{article}

\usepackage[]{emnlp2023}

\usepackage{times}
\usepackage{latexsym}
\usepackage{url}
\usepackage{graphicx}
\usepackage{multirow}
\usepackage{multicol}
\usepackage{lipsum}
\usepackage{subcaption}
\usepackage{subfloat}
\usepackage{amsmath}
\usepackage{caption}
\usepackage{color}
\usepackage{booktabs}
\usepackage{xcolor}
\usepackage{xspace}
\usepackage[T1]{fontenc}

\usepackage[utf8]{inputenc}
\usepackage[normalem]{ulem}
\useunder{\uline}{\ul}{}
\usepackage{microtype}

\title{Understanding Social Structures from Contemporary Literary Fiction using Character Interaction Graph - Half Century Chronology of Influential Bengali Writers}

\author{Nafis Irtiza Tripto\textsuperscript{1,2*},  Mohammed Eunus Ali\textsuperscript{2}\\
        \textsuperscript{1}The Pennsylvania State University, USA\\ 
       \textsuperscript{1}\texttt {nit5154@psu.edu} \\
         \textsuperscript{2}Bangladesh University of Engineering and Technology, Bangladesh   \\ 
          \textsuperscript{2}\texttt {eunus@cse.buet.ac.bd} 
         }

\begin{document}
\maketitle

\begin{abstract}
  
Social structures and real-world incidents often influence contemporary literary fiction. Existing research in literary fiction analysis explains these real-world phenomena through the manual critical analysis of stories. Conventional Natural Language Processing (NLP) methodologies, including sentiment analysis, narrative summarization, and topic modeling, have demonstrated substantial efficacy in analyzing and identifying similarities within fictional works. However, the intricate dynamics of character interactions within fiction necessitate a more nuanced approach that incorporates visualization techniques. Character interaction graphs (or networks) emerge as a highly suitable means for visualization and information retrieval from the realm of fiction. Therefore, we leverage character interaction graphs with NLP-derived features to explore a diverse spectrum of societal inquiries about contemporary culture's impact on the landscape of literary fiction. Our study involves constructing character interaction graphs from fiction, extracting relevant graph features, and exploiting these features to resolve various real-life queries. Experimental evaluation of influential Bengali fiction over half a century demonstrates that character interaction graphs can be highly effective in specific assessments and information retrieval from literary fiction. Our data and codebase are available\footnote{\url{https://cutt.ly/fbMgGEM}}.
  
\end{abstract}
\section{Introduction}
\label{sec-intro}

\begin{figure}[h]
	\centering
	\begin{subfigure}{0.45\columnwidth}
		\centering
		\includegraphics[width=\textwidth]{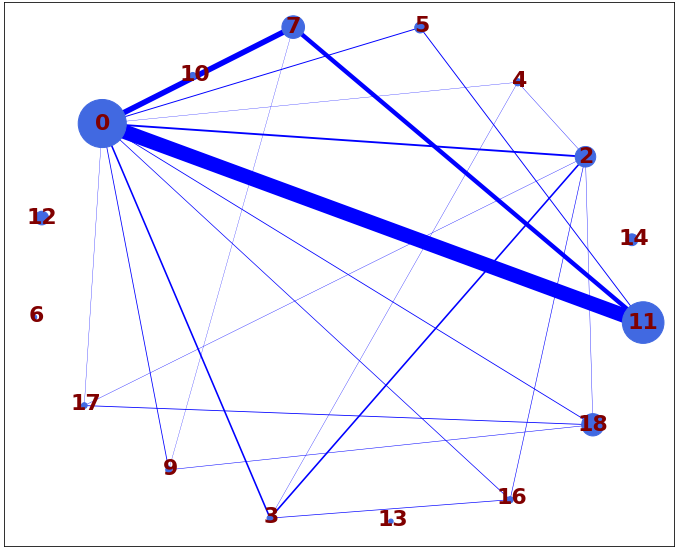}
		\caption{\textit{Shesher Kobita }}
		\label{subfig-long}
	\end{subfigure}%
	\begin{subfigure}{0.45\columnwidth}
		\centering
		\includegraphics[width=\textwidth]{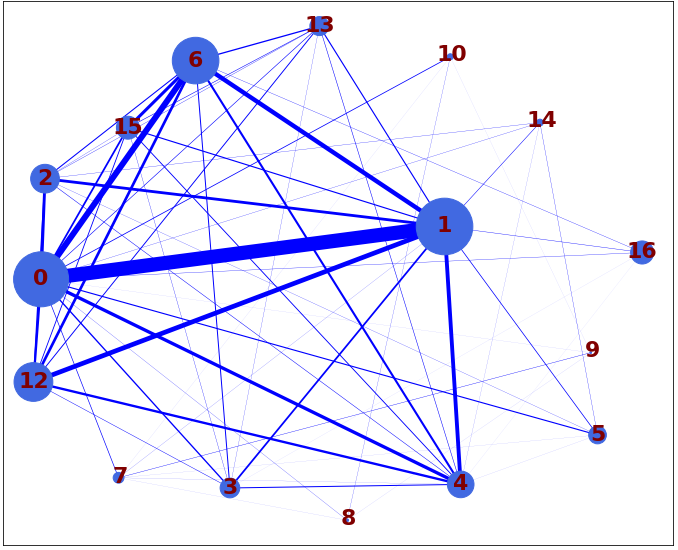}
		\caption{\textit{Gora}}
		\label{subfig-novel}
	\end{subfigure}%
	\caption{Character interaction graph on two novels of Rabindranath Tagore (\textit{(a) The last poem, (b) European}). The bigger the node/thicker the edge is, indicate more weight to the corresponding character/relation. 
	}
	\label{fig-example}
\end{figure} 

\begin{figure*}
    \centering
    \includegraphics[width=1.8\columnwidth]{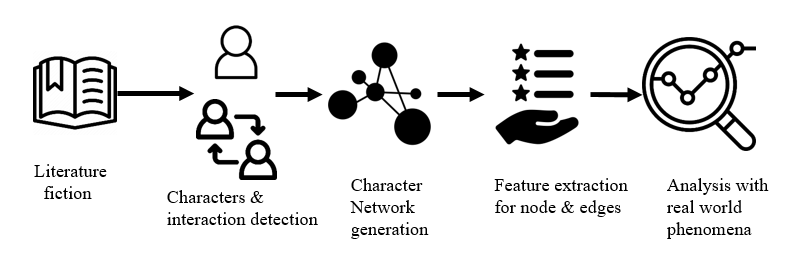}
    \caption{An overview of using character interaction graphs (character networks) for contemporary literary analysis. }
    \label{fig:method}
\end{figure*}

Literary fiction, a reflection of societal values and culture \citep{sadraddinova2019literature}, often employs narrative form to convey its tales. Within these narratives, character interactions drive the plot, constructing personas through their various engagements \cite{min2016network,truby2008anatomy}. Character interaction graphs visually represent these interactions and offer a versatile tool for exploring literary theories and depicting social structures \citep{labatut2019extraction_review}. Beyond this, they prove instrumental in solving diverse literary challenges, from role detection \cite{jung2013emotion} to genre classification \cite{gil2011extraction,ardanay2015clustering,agarwal2021genre} and storyline analysis \cite{weng2007movie}. This paper introduces a data-driven approach that leverages character interaction graphs to elucidate the impact of social structures and contemporary events on literary fiction. Our study delves into the works of influential Bengali writers spanning over half a century, establishing a compelling connection between these real-world influences and the realm of literary fiction.

In our pursuit of understanding the intricate connection between language, history, and literature, we have chosen to focus on Bengali literature. This decision stems from our specialized domain knowledge and the absence of prior quantitative assessments in this area. While a previous study by \citet{muhuri2018bangla_char} visualized character networks in two plays by Rabindranath Tagore, it delved into only limited aspects of character interaction. However, literature mirrors writers' societal perspectives on historical events, gender roles, and more \citep{reynolds1990girls,jarrott2013analysis,white2002historical}. Past research has explored these facets within Bengali literature in addressing the role of women \cite{sen2002woman,chatterjee2009women3,banerjee1989marginalization}, the influence of nationalist movements \cite{majumder2016can}, particular views on religion \cite{quayum2015hindu_muslim,das2012religion}, and the social changes reflected \cite{chaudhuri1971social}. However, these approaches are performed primarily through manual, non-technical analysis. Thus, they tend to overlook significant details in lengthy narratives, leaving the writer's portrayal of their viewpoint through plots and characters unverified. In contrast, we adopt a computational approach, harnessing character interaction graphs' dynamics to unveil the influence of social structures and contemporary events in modern Bengali literature. 

Character interaction graphs, or character networks, are graphical representations derived from a story's narrative, where nodes represent characters, and edges signify their interactions. To illustrate, consider Figure~\ref{fig-example}, which visualizes character interaction graphs from two novels by the renowned Bengali author Rabindranath Tagore. In the first novel, characterized as a romance, a prominent and meaningful connection between the central male and female characters is evident. Conversely, the second novel, with a political theme, introduces a larger ensemble of characters and interactions. Notably, the higher graph density and an increased number of nodes with greater weight highlight the intricate nature of character relationships within political contexts.


Therefore, the primary objective of this paper is to investigate whether character interaction in fiction can depict real-world social structure and perspective from writers. Specifically,  we aim to answer the following research questions (RQs).
\begin{itemize}
\item {\verb|RQ 1|} How various historical events have impacted the character development and prominence of characters in Bengali literature?
\item {\verb|RQ 2|}: To what extent the impact of various age \& gender groups in Bengali society can be inferred through contemporary novels?
\item {\verb|RQ 3|}: Can the presence of different characters and character interaction graph structure be interpreted by the story's context or genre?
\end{itemize}


To answer these questions, we rely on the novels of the three most prominent writers at the beginning of modern Bengali literature (Rabindranath Tagore, Bankim Chandra, Sarat Chandra Chattopadhyay) whose combined literary career span more than a half-century (1865-1935). Additionally, we consider the novels of contemporary Bengali writers Sunil Gangopadhyay and Humayun Ahmed to validate our findings in the modern literary context. Figure \ref{fig:method} provides an overview of our approach. First, we construct character interaction graphs based on character co-occurrence in the narrative. We enrich our analysis by extracting various attributes from nodes and edges, incorporating sentiment \& other NLP features from the story text, and annotating characters with age, gender, role, and other relevant information. Finally, we explore these features for each writer, employ statistical significance tests to affirm our findings and provide a multifaceted evaluation of the results.

 

Our study reveals that historical events like the widow remarriage law in Hindu society (1872) and nationalist movements such as the partition of Bengal (1906) \& Gandhi's non-cooperation movement (1920) substantially impacted character interactions in contemporary literature. Moreover, despite a lower presence of female characters, their collective influence equaled or exceeded that of male characters. Also, the influence of older age groups diminished among writers who had experienced various nationalist movements. Therefore, analyzing character interaction graphs in fiction can provide valuable insights into the social dynamics of specific historical periods.

\section{Related Work}
\label{sec-related_works}



\paragraph{Character interaction graphs:}

Character interaction representations are extensively used in digital humanities to visualize relationships between literary characters. Numerous variations and approaches to character networks exist. For instance, \citet{elson2010extracting} created a network from dialogue interactions in nineteenth-century British novels, where vertices represented characters and edges indicated the frequency and length of their conversations. \citet{elsner2012character} introduced a kernel to measure novel similarity based on characters and their relationships. \citet{ardanay2015clustering} built social networks of characters to represent narrative structures in novels, using EM clustering to group novels by genres and authorship. The primary distinction between different character interaction works lies in character identification, interaction detection, graph creation, and scope of application~\cite{labatut2019extraction_review}. 
Apart from literature, character interaction graph has also gained popularity in other media such as film \citep{cipresso2016computational_movie}, drama \citep{moretti2011network_play}, TV series \citep{weng2007tv_series}, and pop culture.

\paragraph{Social network analysis from character interaction: } 

Character interaction graph has also been utilized to answer various social science questions of contemporary times. \citet{lauzen2005maintaining} discuss the portrayals of different age groups and gender roles in top-grossing Hollywood films. They observe that both older men and women are dramatically underrepresented compared to their representation in real-life. 
Recently \citet{kagan2020using} investigated gender bias in on-screen female characters over the past century using a huge corpus of movie social networks. They discovered a trend of improvement in all aspects of women‘s roles in movies, including a constant rise in the central characters.

The only prior study that focuses on character interaction in
Bengali literature is \citet{muhuri2018bangla_char}. They have extracted
character networks from two plays of Rabindranath Tagore
and proposed a novel idea to analyze the characteristics of
protagonist and antagonist from the influential nodes based
on the complex graph. However, their study does not explain the role of contemporary social set up or gender/age group effect in the character interaction. 
Therefore, our study aims to fill this gap by performing an quantitative assessment in Bengali literature analysis that exploits the character interaction graph to answer these questions.

\section{Methodology and Experiments}
\label{sec-method}

We resort to character interaction graph model to answer our research questions in the context of Bengali literature. Our novel contributions for these tasks are as follows. 

    


\begin{itemize}
    \item We adapt the procedure as discussed below to construct the character interaction graph from story text and character list for Bengali fiction. 
    \item We analyze these graphs from various perspectives and draw the connection to answer our RQs.
    \item We create a novel dataset containing the 63 fictions of five prominent Bengali writers and visualize the character interaction graphs. 
\end{itemize}



\subsection{Character Interaction Graph Generation}

Extracting character interaction graphs from literary text mostly consists of three primary steps: 1) identification of characters, 2) detection of their interactions, and 3) extraction of the interaction graph~\cite{labatut2019extraction_review}.   We have to modify these steps that would apply to our analysis in Bengali fiction. Since stories are collected in chapters, we perform these tasks and create character interaction graphs for each chapter like previous researches~\cite{ardanay2015clustering, agarwal2014frame}. Finally, we combine these chapter-wise graphs to construct the overall story graph.

\paragraph{Character identification:}
Character identification consists of detecting which characters appear in the story and precisely when they appear in the narrative. 
Current Named Entity Recognition (NER) methods~\cite{ chowdhury2018NER,alam2020proposed,mandal2022natural} in the Bengali language are not adequate to find correct character names in the context of literary fiction. 
Hence, our study employs a meticulous approach to character identification that combines automated detection with manual annotation.

For each story, we leverage the BNLP toolkit \citep{sarker2021bnlp} for NER recognition of individuals and cross-verify the character list by briefly reviewing the narrative. We add any missing characters as needed and remove any person names not integral to the narratives.
In cases where a character assumes multiple aliases, we include all the names they adopt, accompanied by using pronouns, particularly for stories narrated in the first person. To identify a character's presence in the story text, we append relevant suffixes and inflections to each name. We assume that if a character's name appears anywhere within the story, they are considered present in that section.


\paragraph{Interaction detection:}

Our approach aligns with prior research that argues that the simple co-occurrence of two characters indicates an interaction \cite{labatut2019extraction_review}. In our study, we adopt the sentence as the fundamental narrative unit and posit that two characters interact when they emerge within the same or nearby sentences. To facilitate this, we employ character occurrence data to partition chapters into smaller segments, subsequently identifying their intersections. We present the detailed methodology in the Appendix.

\paragraph{Graph Generation:}

Our methodology begins by constructing character interaction graphs at the chapter level, subsequently integrating these into a comprehensive story graph (Figure \ref{fig-story_graph} in Appendix). Each chapter's influence on the total story graph is proportionate to the number of sentences it contains. Nodes within the graph represent characters appearing in at least one segment of a given chapter, and edges connect nodes corresponding to characters interacting in at least one segment. We calculate node and edge weights based on various factors, including segment lengths, character appearances, and additional characteristics. Moreover, we incorporate sentiment scores, topic distributions, and supplementary data for both nodes and edges. While we offer a brief overview of node and edge weighting here, we provide comprehensive details on other significant attributes and methodologies in the Appendix.


\textit{Node weight:} 
A character's weight depends on the segment length and the number of times the character is addressed \citep{wolyn2023character_weight}. Also, subsequent segments for a character in a chapter should indicate its higher weight than other characters present in fewer segments. Therefore, we consider a scaling factor $\alpha = 0.1$ as the number of segments increases for a character similar to \citet{seo2013characteristic_weight}. 
Given a character $C$ is present $s_C$ segments in a chapter, length of the segment $i$ is $l_i$ and $C$ is addressed in $l^ \prime_i $ sentences in that segment. If the total chapter length is $L$ and $\beta =0.1$ is the extra weight for the sentences that contain character $C$ \citep{wolyn2023character_weight}, the weight of the corresponding node is defined as 
\[\omega_C = \frac{1}{L} \sum \limits^{s_C}_{i=1} (1+ i \times \alpha) (l_i + \beta \times l^ \prime_i) \].

\textit{Link weight: } We adopt a frequency-based~\cite{elson2010extracting} method to calculate edge weight. Interaction weight between two characters $C_1, C_2$ depends on the number of segments they interact with $s_{C_1C_2}$, segment length $l_i$, number of sentences they are present individually $l^\prime_i$ with scaling weight $\beta$ and number of sentences they are present both $l^{\prime \prime}_i$ with scaling weight $\gamma = 2 \times \beta$. The corresponding weight of the edge is defined as.
\[ \omega_{\langle C_1, C_2 \rangle} = \frac{1}{L} \sum \limits^{s_{\langle C_1, C_2 \rangle}}_{i=1} (1+ i \times \alpha) (l_i + \beta \times l^ \prime_i + \gamma \times l^{\prime \prime}_i ) \]


\subsection{Graph Features Extraction}
Following the methodology of previous works \citep{elsner2012character,elson2010extracting,muhuri2018bangla_char}, our analysis encompasses various attributes, including weight, degree, strength (sum of weights over the edges attached to the node), chapter presence, graph density, and other structural characteristics  related to nodes, edges, and the entire graph. Additionally, we measure the sentiment scores and topic distributions associated with each node. For character nodes, we consider a range of manual attributes, including protagonist status (protagonist/antagonist/regular), gender (male/female), age group, family status (father/mother/uncle/aunt/brother to central characters), religion, social status (poor/wealthy/landlord), all aimed at elucidating connections in line with our research questions. Recognizing the limited availability of age information for most characters, we estimate three distinct age groups, mirroring real-life demographics as closely as possible.

\begin{itemize}
\item {\verb|Age group A1: <20 year|}:  This group mostly consists of children and adolescents. 
\item {\verb|Age group A2: 20-40 year|}: Young adults and early middle-aged persons who serve as the current generation in story. 
\item {\verb|Age group A3: >40 year|}: Older people. They usually play the role of the previous generation of young people (A2 group). 
\end{itemize}

\subsection{Dataset}
Our primary focus centers on three eminent writers, namely, Bankim Chandra (BC), Rabindranath Tagore (RT), and Sarat Chandra Chattopadhyay (SC), who belong to the early period of modern Bengali literature. Our study concentrates on their fiction works, particularly novels, as they possess distinctive attributes that set them apart from non-fictional writings \cite{labatut2019extraction_review}. We analyze a selection of their novels, spanning various genres such as historical, romantic, social, and political. Additionally, for comparative purposes with contemporary literature, we delve into the works of two renowned modern writers: Sunil Gangopadhyay (SG) and Humayun Ahmed (HM). Throughout the remainder of this paper, we will employ the authors' first names or abbreviated forms to represent them.

Table \ref{tab:dataset} presents a comprehensive overview of our dataset. The novels were purchased in ebook format, and the text underwent the procedures detailed earlier to create and extract character interaction graphs. To validate the character lists and assign attributes like age, gender, and other status to the characters, we engaged two annotators well-versed in the novels' contents. As a token of appreciation for their contributions, the annotators received gift cards equivalent to \$20.00 each. We cannot release the original text due to copyright constraints. However, we have made our generated character interaction graphs, extracted features, and code-base publicly accessible \footnote{\url{https://cutt.ly/fbMgGEM}}.

\begin{table}[h]
\begin{tabular}{llc}
\toprule
Author                                                                     & Career    & \# novel \\ \midrule
Bankim Chandra (BC)                                                        & 1865-1885 & 12 \\
Rabindranath Tagore (RT)                                                   & 1883-1935 & 11 \\
\begin{tabular}[c]{@{}l@{}}Sarat Chandra \\ Chattopadhay (SC)\end{tabular} & 1907-1940 & 16 \\
Humayun Ahmed (HM)                                                         & 1970-2011 & 15 \\
Sunil Gangopadhyay (SG)                                                    & 1965-2012 & 14 \\ \bottomrule
\end{tabular}
\caption{Overview of dataset}
\label{tab:dataset}
\end{table}

\begin{table*}[h]
\centering
\begin{tabular}{c|cc|cc||cc|cc|cc}

\toprule
Writer                      & Male & $\omega_{\langle M \rangle}$ & Female & $\omega_{\langle F \rangle}$  & A1 & $\omega_{\langle A1 \rangle}$ & A2 & $\omega_{\langle A2 \rangle}$ & A3 & $\omega_{\langle A3 \rangle}$ \\
\midrule
BC             & 0.6086                    & { \ul 0.4273}                            & 0.3913                       & { \ul 0.5727}                            & 0.0729                   & 0.103                          & 0.5620                   & 0.7327                         & 0.3649                   & 0.1643                         \\
RT        & 0.6410                     & 0.5773                            & 0.3589                       & 0.4227                            & 0.0172                   & 0.0293                         & 0.6982                   & 0.8497                         & 0.2844                    & 0.121                          \\
SC & 0.6936                     & { \ul 0.5457}                            & 0.3063                        & { \ul 0.4543}                            & 0.0990                    & 0.103                          &  0.6081                    & { \ul 0.7327}                         & 0.2927                    & 0.1643                         \\
HM             & 0.6726                      & 0.6669                            & 0.3273                        & 0.3331                            & 0.0778                    & 0.0467                         & 0.5628                   & 0.6234                         & 0.3592                    & 0.3299                         \\
SG         & 0.7058                     & 0.7573                            & 0.2941                       & 0.2427                            & 0.0441                   & 0.0292                         & 0.7794                  & 0.9047                         & 0.1764                   & 0.0661         \\
\bottomrule
\end{tabular}
\caption{Age \& gender-wise proportion and aggregate weight (normalized form) for each writer. The {\ul underlined} value represents that a statistical significance was found between male and female characters in weight.}
\label{tab-age_gender_weight}
\end{table*}

\section{Results and Findings}
\label{sec-results}

Character interaction graphs, enriched with various attributes, including descriptive details, sentiment scores, and topic information, offer a unique perspective on character presence and interactions within a story. This approach outperforms traditional manual analysis in terms of efficiency and effectiveness. For instance, by examining factors like character count, weight, degree, sentiment, and protagonist status, especially in the context of different age or gender groups, we can determine which groups exert the most influence in fictional works. Furthermore, graph topology analysis from stories of distinct contexts/genres allows us to validate the representation of social structures in fiction and assess the impact of contemporary events on these narratives.

This section showcases our primary discoveries from various angles, paving the way for exploring their links to our research questions in the subsequent section. We provide concise insights into the roles played by different age and gender groups, protagonist attributes, and variations in graph structures. To assess the significance of these distinctions based on gender or age, we employ the independent two-sample t-test \citep{keselman2004new_t-test}. This statistical test is chosen for its suitability in cases where the two samples are independent and originate from populations with roughly normal distributions \citep{manfei2017differences_t-test}.

\paragraph{Age and gender distribution:}
How age and gender are depicted in popular media is an interesting area of study~\cite{lauzen2005maintaining,kagan2020using} and can portray  writers' perspective on social structure ~\cite{bilston2004awkward}. Table~\ref{tab-age_gender_weight} demonstrates the proportion of different age \&  gender groups and the mean (over stories) of aggregated weight $\omega$ across various groups in all writers' works.  Notably, male characters are more prevalent than female characters across all writers, aligning with prior research findings in different media contexts \cite{lauzen2005maintaining,kagan2020using}. Additionally, age group A2 appears more frequently than A1 and A3 for all writers.

\begin{figure*}[t]
	\centering
	\begin{subfigure}{0.66\columnwidth}
		\centering
		\includegraphics[height=2.8cm]{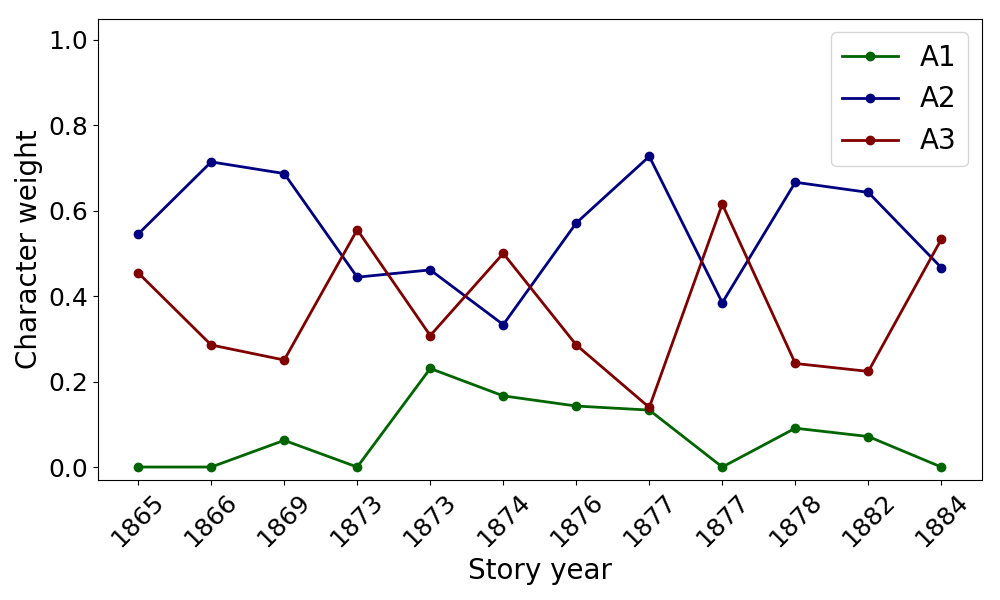}
		\caption{Bankim Chandra}
	\end{subfigure}%
	\begin{subfigure}{0.66\columnwidth}
		\centering
		\includegraphics[height=2.8cm]{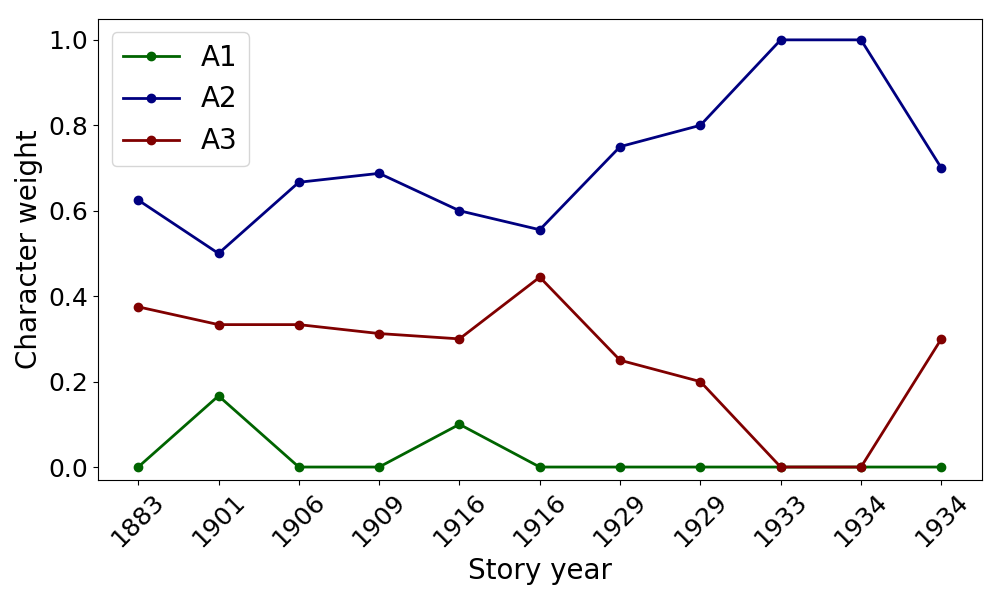}
		\caption{Rabindranath Tagore}
	\end{subfigure}%
		\begin{subfigure}{0.66\columnwidth}
		\centering
		\includegraphics[height=2.8cm]{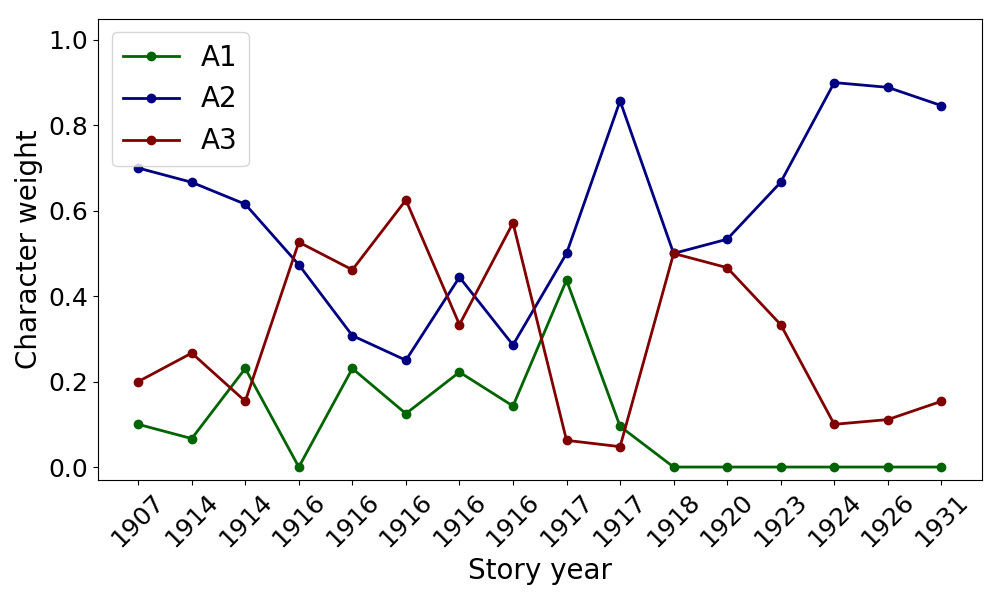}
		\caption{Sarat Chandra Chattopadhyay}
	\end{subfigure}%
	\caption{Distribution of different age group over time for different authors. 
	}
	\label{fig-age_group}
\end{figure*}  

Bankim and Humayun's fiction tends to feature more aged characters from the A3 group. Bankim's novels evoke a feudalistic societal structure \cite{chaudhuri1971social}, which is reflected in the prevalence of older characters. Humayun, on the other hand, focuses on middle-class family struggles in contemporary settings, hence the higher representation of older characters \cite{mamun2014humayun}. Sarat's fiction resembles Bankim's in terms of the presence of A1 characters, often in central roles due to the prevalence of early marriages in their context \cite{chaudhuri1971social}. However, Rabindranath's contemporary fiction, characterized by characters from the upper-middle class with liberal education \cite{park2012social}, sees fewer A1 characters and a relatively higher presence of A2 characters than Bankim and Sarat's narratives.  An intriguing observation emerges as the female character percentage does not increase for the fiction of modern-day writers, which shows an exception from existing studies in other media \cite{kagan2020using}. Notably, the weight attributed to female characters in modern Bengali literature surpasses their relatively lower representation, with statistical significance observed for Bankim and Sarat; this phenomenon is in line with the historical trend of early Bengali literature where women played central roles in plot development \cite{chatterjee2009women3,sen2002woman}, but contemporary writers exhibit gender-neutral character weight distribution, reflecting their distinct narrative priorities.


In Figure \ref{fig-age_group}, we present the evolving proportions of different age groups in the works of three previous writers. A notable surge in the A1 age group occurs in Bankim's writings between 1873 and 1877, coinciding with the enactment of the widow remarriage law in 1872 \citep{mukherjee1985women_bankim} (as listed in Table \ref{tab-events} in Appendix). In 1916, during the economic turmoil resulting from World War I, Sarat's fiction prominently featured the A3 age group revolving around rural society struggles \citep{dutt1981novelist_sarat}. Subsequently, from approximately 1918 onwards, Rabindranath and Sarat witnessed a significant rise in the A2 age group's presence, alongside a decline in the A3 age group. This shift aligns with the influence of nationalist movements and non-cooperation activities \citep{gupta2016rabindranath_works}, prompting their fiction to transition from romantic and conventional social issues to more politically and socially crisis-oriented narratives, characterized by an increased presence of A2 characters and the near absence of A1 characters in their writings during this period.

\begin{figure*}
    \centering
    \includegraphics[scale=0.45]{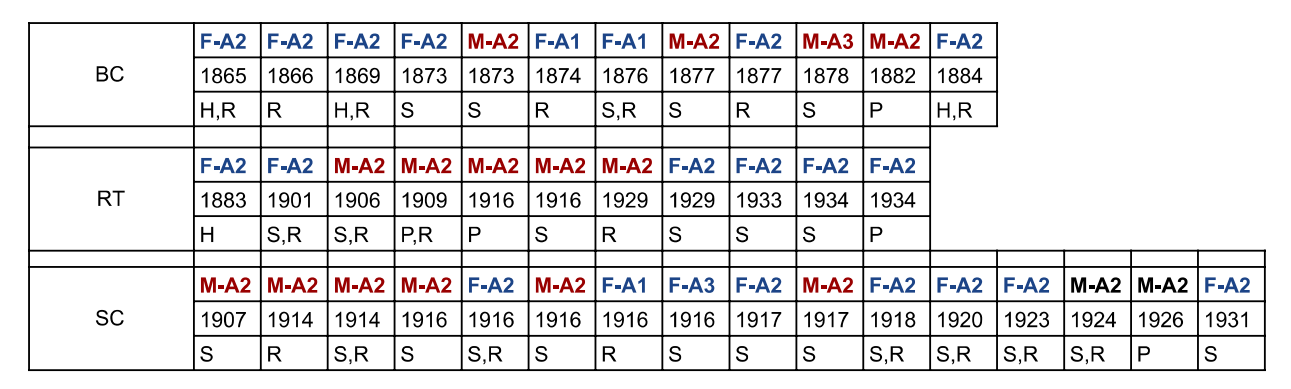}
    \caption{Protagonist and genre information in chronological order (S:Social, P:Political, R:Romantic, H:Historical). }
    \label{fig:pro_genre_year}
\end{figure*}

\paragraph{Protagonist characteristics:}

\begin{table*}[h]
\centering
\begin{tabular}{l|ll|ll|ll|ll}
\toprule
Writer & M (\%) & F (\%) & $\overline{\omega}_{\langle M \rangle}$ & $\overline{\omega}_{\langle F \rangle}$ & $\overline{D}_{\langle M \rangle}$ & $\overline{D}_{\langle F \rangle}$ & $\overline{S}_{\langle M \rangle}$ & $\overline{S}_{\langle F \rangle}$ \\
\midrule
BC    & {\ul 0.6153}                     & {\ul 0.3847}                     & {\ul 0.5917}                   & {\ul 0.8724}                   & 9.22                     & 9                        & {\ul -0.2384 }                 & {\ul -0.4113}                  \\
RT    & 0.444                      & 0.556                      & {\ul 0.244}                    & {\ul 0.8974}                   & 9.2                      & 7.85                     & {\ul 0.1482}                  & {\ul -0.116}                    \\
SC    & 0.6                        & 0.4                        & 0.5111                   & 0.5091                   & {\ul 8}                        & {\ul 4.4}                      & -0.1833                  & -0.2109       \\
\bottomrule
\end{tabular}
\caption{Protagonist Characteristics for different writers based on gender (male/female):  weight ($\overline{\omega}$), degree ($\overline{D})$, and sentiment score ($\overline{S})$.  {\ul Underlined} value indicates  statistical significance was found in that criteria.}
\label{tab-protagonist_char}
\end{table*}

Figure \ref{fig:pro_genre_year} provides insights into protagonist information, while Table \ref{tab-protagonist_char} delves into the characteristics of these protagonists, revealing that most of them belong to the young A2 age group. Bankim's narratives include some A1 female protagonists following the widow remarriage law, and Sarat features one romantic novel with a female A1 group protagonist. Bankim and Sarat also have a social novel with an A3 group protagonist. In his urban-centric plots, Rabindranath portrays all his protagonists as A2 age group characters. Notably, all of Bankim's historical and romantic novels feature female protagonists, whereas both Rabindranath (in \textit{"Shehser Kabita"}) and Sarat (in \textit{"Devdas"}) present romantic novels with male protagonists. Rabindranath's earlier social and political novels in the early twentieth century predominantly revolved around male protagonists, while his later works in these genres incorporate female leads, potentially influenced by the active participation of women in various nationalist movements in the 1920s \cite{sarkar1987nationalist,sen2002woman}.

Despite having lower connectivity, female protagonists exhibit relatively higher weight in the narratives of all writers. They carry a slightly negative emotional sentiment, often signaling tragic endings in the stories \citep{kaviraj1995bankim_works}. Additionally, their topic distribution tends to be concentrated on specific themes, such as social and family matters, while male protagonists feature a more diverse range of topics in their narratives.

\begin{table}[h]
\centering
\begin{tabular}{llll}
\toprule
Author                      & \# Node & Density & \# Edge \\
\midrule
BC            & 11.6923 & 0.4952  & 32.2308 \\
RT        & 10      & 0.4565  & 19      \\
SC & 14.4375 & 0.363   & 37.25  \\
\bottomrule
\end{tabular}
\caption{Graph structure property for different auhtors}
\label{tab-deg_density_count}
\end{table}

\begin{figure}[h]
    \centering
   	\includegraphics[height=3.5cm]{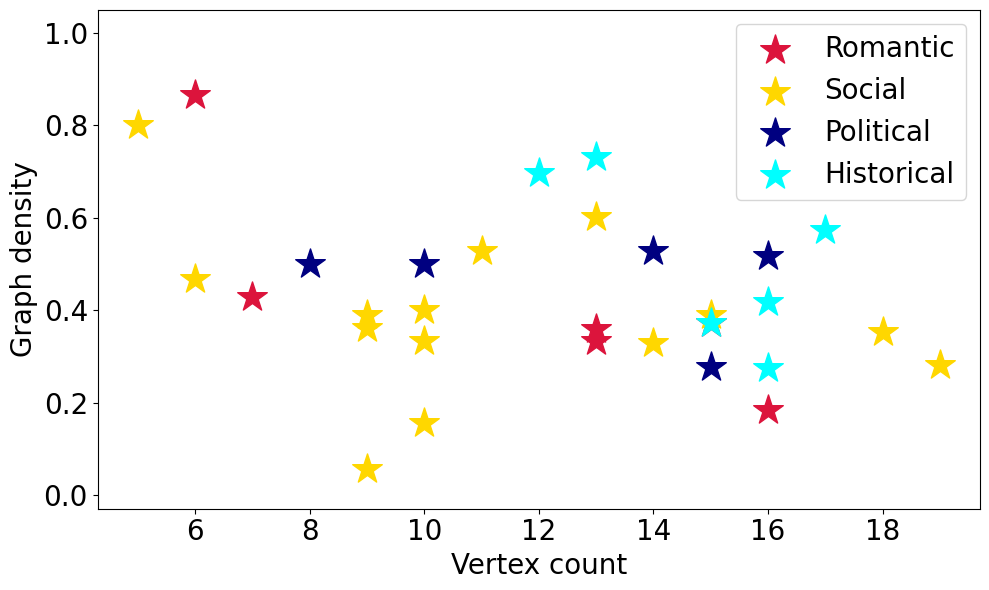}
		\caption{Node count \& graph density}
		\label{subfig-node_density}
\end{figure}

\paragraph{Variation in graph structure:}
 First, we present the average count of node, edge and graph density for all writers in Table \ref{tab-deg_density_count}. 
 Rabindranath's portrayal of the higher middle-class, educated urban society \citep{collins2008rabindranath, sen2002woman, gupta2016rabindranath_works} is characterized by a compact social structure with fewer nodes, in contrast to Sarat's depiction of rural society \citep{dutt1981novelist_sarat}, which involves more characters but with a smaller density. Furthermore, Figure \ref{subfig-node_density} illustrates the relationship between graph density and node count across different genres. Romantic novels exhibit either small, dense networks (fewer nodes but higher density) or large, sparse networks (more nodes but lower density). Historical fiction typically features many characters, while political novels maintain a high graph density even as the character count increases.

\section{Discussion}
\label{sec-discussion}

Based on our key findings in the previous section,  we answer our research questions and validate our assumptions in this section. 

\subsection{Influence of Real-life Events }

To investigate the influence of historical and social events on contemporary fiction, we compile a list of noteworthy national events during our study period (see Table \ref{tab-events} in the Appendix). Specifically, the widow remarriage law in Hindu society (1872) substantially impacts Bankim's contemporary novels, as elaborated below. Additionally, during the nineteenth century, various nationalist movements inspired Rabindranath and Sarat, leading them to produce several social and political novels, further detailed in the Appendix.

\begin{figure}[h]
	\centering
	\begin{subfigure}{0.45\columnwidth}
		\centering
		\includegraphics[width=\textwidth]{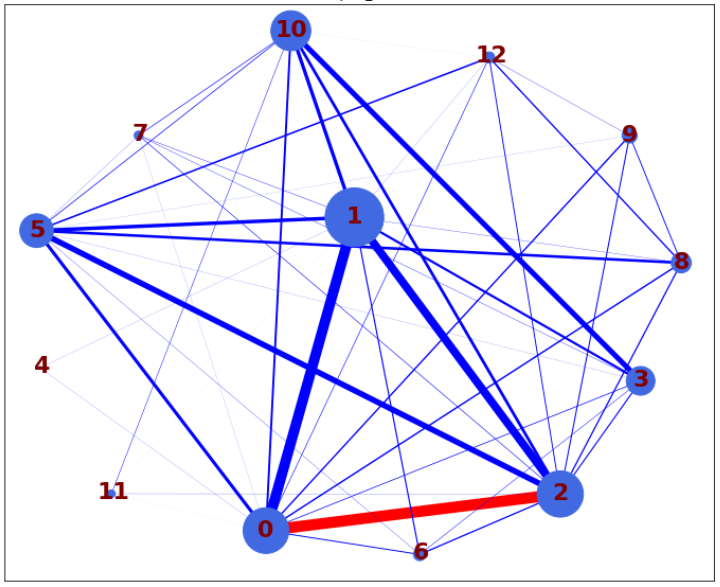}
		\caption{\textit{Bisabrksa (The Poison Tree, 1872)}}
		\label{subfig-brishbikkha}
	\end{subfigure}%
	\begin{subfigure}{0.45\columnwidth}
		\centering
		\includegraphics[width=\textwidth]{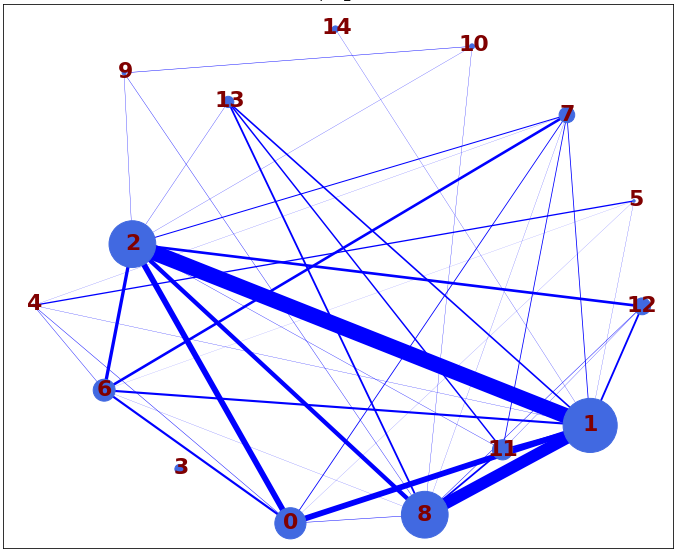}
		\caption{\textit{Krishnakanter Will (The Will of Krishnakanta, 1878)}}
		\label{subfig-krisnokanto}
	\end{subfigure}%
	\caption{Character interaction graph for two novels of Bankim. Protagonist  of both stories are widow.
	}
	\label{fig-widow}
\end{figure}  

\paragraph{Impact of widow remarriage law:}

The Brahma Marriage Act of 1872, which lifted the ban on widow remarriage, notably influenced Bankim's works. Bankim had previously addressed this issue in various non-fiction writings \cite{kaviraj1995bankim_works}. Following the passage of the Remarriage Act, there was a significant increase in the presence of female A1 characters in Bankim's writings. This shift is particularly evident in his novels, \textit{"Bisabrksa"} (The Poison Tree, 1872) and \textit{"Krishnakanter Will"} (The Will of Krishnakanta, 1878).

Bankim's opposition to the widow remarriage law is evident in his novels, where he incorporates widows into complex relationship triangles with married men and their lawful wives (1-0-2 in \ref{subfig-brishbikkha} and 2-1-8 in \ref{subfig-krisnokanto}). These relationship triangles are visually represented in the character interaction graphs, accompanied by an overall negative sentiment. Moreover, previous studies have confirmed the predominantly negative outcomes of these stories \citep{kaviraj1995bankim_works}. However, two of Bankim's novels during this period, \textit{"Yugalanguriya"} (1874) and \textit{"Radharani"} (1876), feature female A1 protagonists who are not widows, and these graphs do not exhibit such relationship triangles.



\subsection{Influence of Age and Gender Group}

In earlier times, despite a lower ratio of female characters, their weight in the narrative was significant due to stories centered around women and their societal roles. However, contemporary writers no longer consistently emphasize increased weight for female characters. While political novels typically exclude female characters from influential roles, the situation evolves with the involvement of women in nationalist movements. The prevalence of A2 age group characters aligns with societal norms, while the appearance of A1 female characters as central figures mirrors early marriage practices influenced by real-life events. Post-1916, social and nationalist movements reshaped novels, shifting them towards socio-political themes and significantly increasing A2 group representation while other groups diminished in importance.

\subsection{Interpretation of Graph from Context}

Finally, we assess whether the graph's topological structure and character presence can be inferred from the context or genre of the fiction. Bankim's novels reflect a feudalistic social structure with landlords and kings, thus incorporating more aged characters than other writers. Rabindranath's urban-centric plots feature upper-middle-class educated characters, predominantly young, with a greater emphasis on female protagonists but fewer noticeable female A1 group characters compared to Bankim or Sarat. Sarat's rural settings include fewer female characters, yet their presence, connectivity, and weight are more pronounced. Minor characters in Rabindranath's fiction have significantly lower node counts and edge weights than Sarat's, reflecting the urban setting's fewer characters and interactions than the rural context.

Similarly, genre shapes the character presence and graph structure in fiction. Romantic novels feature female and male protagonists, leading to densely or sparsely connected networks. Political novels exhibit higher graph density regardless of node count. Historical novels tend to have more nodes, reflecting their expansive nature.

\section{Conclusion}
\label{sec-conclusion}

This paper presents an exploration of social structures within contemporary Bengali literature. We employ character interaction graphs to model the works of prominent Bengali writers spanning over half a century, extracting pertinent features. Our analysis rigorously addresses three pivotal research questions regarding the influence of social structures in Bengali literary fiction. Our findings substantiate the profound impact of historical events, such as the widow remarriage act and nationalist movements, on contemporary literary works. Notably, our study unveils the substantial significance accorded to female characters despite their relatively lower prevalence. By providing visualization and quantitative assessment tools for analyzing influential fiction, our research empowers modern researchers to engage in critical literary analysis.

\section*{Limitations}

Our study has certain limitations that warrant acknowledgment. Given the challenges of working with a low-resource language, our dataset is limited to five writers. The manual annotation of characters and attributes requires enormous effort and detailed knowledge of these novels. Some characters could be missing in our character interaction graphs due to unwanted annotation errors, although these are predominantly minor characters with minimal impact on our analysis quality. We have opted for static graphs (story-wise) in our analysis to specifically examine the influence of contemporary events and character group presence in fiction. Future research avenues could explore the dynamics of character interaction, sentiment, and weight changes throughout the narrative, requiring a separate study. Our future plans also expand our dataset to encompass more writers and diverse chronological periods. We intend to incorporate previously unexplored character attributes, such as religion and economic status, to offer multifaceted insights into our analysis.

\section*{Ethics Statement}

While the ultimate goal of this study is to investigate social structures represented in contemporary Bengali literature through character interaction graphs, we acknowledge the potential for these graphs to unveil sensitive connections such as gender or religious issues that may not have been the writers' original intent. While we support our findings with analyses validated by prior research on the writers' works, it is essential to recognize the possibility of some conclusions being subject to interpretation. Nonetheless, our research contributes valuable visualization and quantitative assessment tools, which can facilitate researchers in conducting rigorous literary analysis with greater ease.

\bibliography{main}
\bibliographystyle{acl_natbib}



\clearpage
\appendix

\begin{figure*}[h]
    
  \centering
  \includegraphics[width=\linewidth]{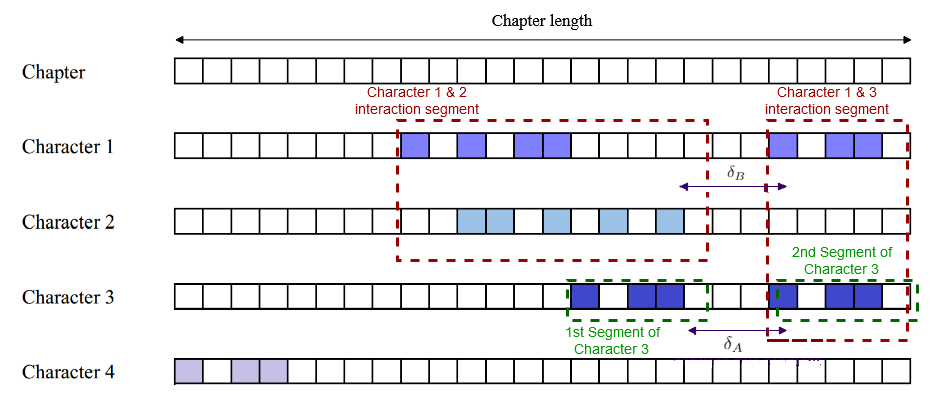}
  \caption{Interaction of different characters in a sample story chapter. Each sentence is denoted as a cell. A filled cell indicates that specific character is appeared in this sentence. Segement 1 and 2 of Character 3 are considered seperate segments because their distance is greater than $\delta_A$. Similarly, segment 2 of character 1 and segment 3 of character 2 do not belong to the same plot since their distance is greater than $\delta_B$. }
  \label{fig-interaction}
\end{figure*}

\begin{figure*}[h]
    \centering
    \includegraphics[width=\linewidth]{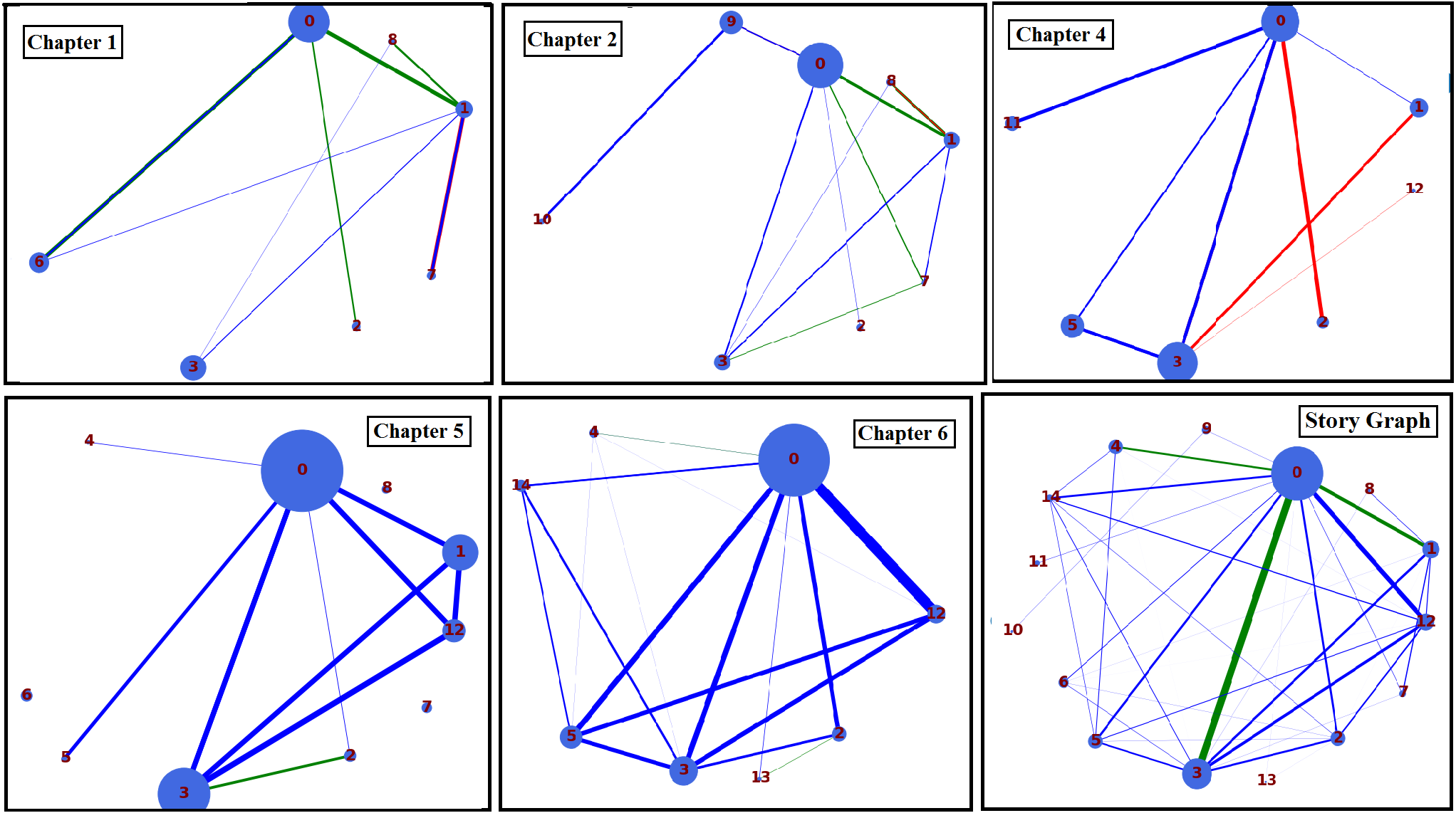}
    \caption{Generation of story graph for novel  \textit{(Devi Chowdhurani 1884)} by Bankim Chandra from corresponding chapter graphs. The final story graph include all nodes and edges that are present at any of the chapter-wise graphs. The node and edge weights in final are computed from the weighted average of these graphs. Blue, green and red edges indicate neutral, positive and negative  sentiment respectively. }
    \label{fig-story_graph}
\end{figure*}

\section{Character Interaction Detection}
Figure \ref{fig-interaction} demonstrates a sample instance of character interact identification procedure. We briefly discuss the interaction detection steps as follows.

\begin{enumerate}
    \item For each character $C$, we utilize the character occurrence to split the chapter into smaller segments. Now, two sentences $i$ and $j$, where the character $C$ is present, will belong to the same segment if $|i-j|< \delta_A$. Here, $\delta_A$ denotes the threshold for intra-character sentence-wise distance. Also, a segment will form only if the number of sentences where the $C$ is present is greater than the minimum appearance threshold value $\delta_C$  in that chapter. Each segment will be denoted as $S := \langle s, f \rangle$, where $s$ and $f$ indicate the starting and finishing sentence of that segment. 
    
    \item Two characters $C_1$ and $C_2$ will have common a segment if any individual segment of $C_1$ overlap with another individual segment of $C_2$. More specifically, characters $C_1$ and $C_2$ will have intersection if $\langle s_1 - \delta_B , f_1 + \delta_B \rangle$ and $\langle s_2 - \delta_B , f_2 + \delta_B \rangle$ overlaps, where $\delta_B$ is the inter-character sentence-wise distance threshold.
    
    \item We utilize these character and interaction segments to calculate weight, sentiment score and other attributes for nodes and edges in character interaction graph. 
\end{enumerate}

The values of $\delta_A$, $\delta_B$, $\delta_C$ are automatically determined from the chapter length and characters count for each chapter of a story. We combine the chapter-level graphs to create the ultimate story graph (Figure \ref{fig-story_graph} as example), which is the basis for our analysis.

\section{Different Character Interaction Graph Attributes}
\paragraph{Sentiment score:}
Existing sentiment analysis approaches in Bengali primarily target specific domains, such as social media platforms like Twitter \cite{chowdhury2014twitter} or YouTube comments \cite{tripto2018detecting}, and are ill-suited for literary text analysis. To address this limitation, we employ a more versatile Bangla-BERT model \citep{kowsher2022bangla_bert} for sentiment analysis, augmented by Bengali SentiWordNet \cite{das2010sentiwordnet} and WordNet affect \cite{das2010emotion}. Sentiment analysis is conducted at the character and interaction levels based on the segments they occupy within a chapter. Initially, we compute the sentiment of individual sentences in the chapter and then standardize the sentiment scores across segments to ensure consistency. Character sentiment (node) and interaction sentiment (edge) are subsequently determined from the average sentiment score of the associated segments.

\paragraph{Topic distribution}
In parallel with our sentiment analysis, we also assign topic information to characters based on their presence within story segments. To achieve this, we employ Latent Dirichlet Allocation (LDA) \cite{blei2003latent} to generate a topic model specific to each writer, utilizing a manually determined number of topics (t=20) with general annotations of the topic name for improved interpretability. In the preprocessing phase, we eliminate stopwords, character names, and commonly used verbs to enhance the granularity of the topic distribution \cite{mehrotra2013improving_topic}. Character-specific topic distributions are established by computing the weighted average of topic scores across all segments featuring the character.

\paragraph{Character importance}
 We also calculate the importance of each character in their interaction as an edge attribute.  Given two characters $C_1$, $C_2$, their segment length be $l_1$ $l_2$ respectively and l is their overlapping length. Then the importance of character  $C_1$, $C_2$ in their corresponding link $\langle C_1, C_2 \rangle$ is defined as. 
\[ \Phi_{C_1} = \frac{l}{l_1} + \frac{\# \; times \; C_1  \; addressed}{l} \]
\[\Phi_{C_2} = \frac{l}{l_2} + \frac{\# \; times \; C_2  \; addressed}{l}\]

\paragraph{Other attributes} For each character and interaction,  we also maintain their sequence (specific position of associated segments in the chapter), total appearance, and segment count.


\begin{table*}[t]
\centering

\setlength{\tabcolsep}{2pt}
\begin{tabular}{|lll|lll|lll|}
\hline
\multicolumn{3}{|c|}{Bankim Chandra}                                                                                                                                        & \multicolumn{3}{c|}{Rabindranath Tagore}                                                                                                                                    & \multicolumn{3}{c|}{Sarat Chandra Chattopadhyay}                                                                                                                      \\ \hline
Story name                                                          & Genre                                                          & P                                    & Story name                                                           & Genre                                                         & P                                    & Story name                                                       & Genre                                                       & P                                    \\ 
\hline
\begin{tabular}[c]{@{}l@{}}Durgeshnan\\ dini(1865)\end{tabular}     & \begin{tabular}[c]{@{}l@{}}Historical \\ Romantic\end{tabular} & {\color[HTML]{00009B} \textbf{F-A2}} & \begin{tabular}[c]{@{}l@{}}Bou Thakuran\\ ir Haat(1883)\end{tabular} & Historical                                                    & {\color[HTML]{00009B} \textbf{F-A2}} & \begin{tabular}[c]{@{}l@{}}Borodidi\\ (1907)\end{tabular}        & Social                                                      & {\color[HTML]{9A0000} \textbf{M-A2}} \\ \hline
\begin{tabular}[c]{@{}l@{}}Kapalakun\\ dala(1866)\end{tabular}      & {\color[HTML]{F56B00} Romantic}                                & {\color[HTML]{00009B} \textbf{F-A2}} & \begin{tabular}[c]{@{}l@{}}Chokher Bali\\ (1901)\end{tabular}        & \begin{tabular}[c]{@{}l@{}}Social,\\ Romantic\end{tabular}    & {\color[HTML]{00009B} \textbf{F-A2}} & \begin{tabular}[c]{@{}l@{}}Devdas\\ (1914)\end{tabular}          & {\color[HTML]{F56B00} Romantic}                             & {\color[HTML]{9A0000} \textbf{M-A2}} \\ \hline
\begin{tabular}[c]{@{}l@{}}Mrinalini\\ (1869)\end{tabular}          & \begin{tabular}[c]{@{}l@{}}Historical \\ Romantic\end{tabular} & {\color[HTML]{00009B} \textbf{F-A2}} & \begin{tabular}[c]{@{}l@{}}Noukadubi\\ (1906)\end{tabular}           & \begin{tabular}[c]{@{}l@{}}Social, \\ Romantic\end{tabular}   & {\color[HTML]{9A0000} \textbf{M-A2}} & \begin{tabular}[c]{@{}l@{}}Panditama\\  sai(1914)\end{tabular}    & \begin{tabular}[c]{@{}l@{}}Social,\\ Romantic\end{tabular}  & {\color[HTML]{9A0000} \textbf{M-A2}} \\ \hline
Indira(1873)                                                        & Social                                                         & {\color[HTML]{00009B} \textbf{F-A2}} & Gora(1909)                                                           & \begin{tabular}[c]{@{}l@{}}Political,\\ Romantic\end{tabular} & {\color[HTML]{9A0000} \textbf{M-A2}} & \begin{tabular}[c]{@{}l@{}}Pallisamaj\\ (1916)\end{tabular}      & Social                                                      & {\color[HTML]{9A0000} \textbf{M-A2}} \\  \hline
\begin{tabular}[c]{@{}l@{}}Bisabrksa\\ (1873)\end{tabular}          & Social                                                         & {\color[HTML]{963400} \textbf{M-A2}} & \begin{tabular}[c]{@{}l@{}}Ghare Baire\\ (1916)\end{tabular}         & {\color[HTML]{32CB00} Political}                              & {\color[HTML]{9A0000} \textbf{M-A2}} & \begin{tabular}[c]{@{}l@{}}Chandranath\\ (1916)\end{tabular}     & \begin{tabular}[c]{@{}l@{}}Social, \\ Romantic\end{tabular} & {\color[HTML]{00009B} \textbf{F-A2}} \\  \hline
\begin{tabular}[c]{@{}l@{}}Yugalangu\\ riya(1874)\end{tabular}      & {\color[HTML]{F56B00} Romantic}                                & {\color[HTML]{00009B} \textbf{F-A1}} & \begin{tabular}[c]{@{}l@{}}Chaturanga\\ (1916)\end{tabular}          & Social                                                        & {\color[HTML]{9A0000} \textbf{M-A2}} & \begin{tabular}[c]{@{}l@{}}Baikunther \\ Will(1916)\end{tabular} & Social                                                      & {\color[HTML]{9A0000} \textbf{M-A2}} \\  \hline
\begin{tabular}[c]{@{}l@{}}Radharani\\ (1876)\end{tabular}          & \begin{tabular}[c]{@{}l@{}}Social, \\ Romantic\end{tabular}    & {\color[HTML]{00009B} \textbf{F-A1}} & \begin{tabular}[c]{@{}l@{}}Shesher\\ Kabita (1929)\end{tabular}      & {\color[HTML]{F56B00} Romantic}                               & {\color[HTML]{9A0000} \textbf{M-A2}} & \begin{tabular}[c]{@{}l@{}}Parinita\\ (1916)\end{tabular}        & {\color[HTML]{F56B00} Romantic}                             & {\color[HTML]{00009B} \textbf{F-A1}} \\  \hline
\begin{tabular}[c]{@{}l@{}}Chandra\\ sekhar(1877)\end{tabular}      & Social                                                         & {\color[HTML]{9A0000} \textbf{M-A2}} & Jogajog(1929)                                                        & Social                                                        & {\color[HTML]{00009B} \textbf{F-A2}} & \begin{tabular}[c]{@{}l@{}}Araksaniya\\ (1916)\end{tabular}      & Social                                                      & {\color[HTML]{00009B} \textbf{F-A3}} \\  \hline
Rajani(1877)                                                        & {\color[HTML]{F56B00} Romantic}                                & {\color[HTML]{00009B} \textbf{F-A2}} & \begin{tabular}[c]{@{}l@{}}Dui bon\\ (1933)\end{tabular}             & Social                                                        & {\color[HTML]{00009B} \textbf{F-A2}} & \begin{tabular}[c]{@{}l@{}}Niskriti\\ (1917)\end{tabular}        & Social                                                      & {\color[HTML]{00009B} \textbf{F-A2}} \\ \hline
\begin{tabular}[c]{@{}l@{}}Krishnakanter \\ Will(1878)\end{tabular} & Social                                                         & {\color[HTML]{9A0000} \textbf{M-A3}} & \begin{tabular}[c]{@{}l@{}}Malancha\\ (1934)\end{tabular}            & Social                                                        & {\color[HTML]{00009B} \textbf{F-A2}} & \begin{tabular}[c]{@{}l@{}}Charitrohin\\ (1917)\end{tabular}     & Social                                                      & {\color[HTML]{9A0000} \textbf{M-A2}} \\  \hline
\begin{tabular}[c]{@{}l@{}}Anandamath\\ (1882)\end{tabular}         & {\color[HTML]{009901} Political}                               & {\color[HTML]{9A0000} \textbf{M-A2}} & \begin{tabular}[c]{@{}l@{}}Char Odhhay\\ (1934)\end{tabular}         & {\color[HTML]{009901} Political}                              & {\color[HTML]{00009B} \textbf{F-A2}} & Datta(1918)                                                      & \begin{tabular}[c]{@{}l@{}}Social, \\ Romantic\end{tabular} & {\color[HTML]{00009B} \textbf{F-A2}} \\  \hline
\begin{tabular}[c]{@{}l@{}}Devi Chaudhu\\ rani(1884)\end{tabular}   & \begin{tabular}[c]{@{}l@{}}Historical\\ Romantic\end{tabular}  & {\color[HTML]{00009B} \textbf{F-A2}} &                                                                      &                                                               &                                      & \begin{tabular}[c]{@{}l@{}}Grihodaho\\ (1920)\end{tabular}       & \begin{tabular}[c]{@{}l@{}}Social, \\ Romantic\end{tabular} & {\color[HTML]{00009B} \textbf{F-A2}} \\ \cline{7-9}
                                                                    &                                                                &                                      &                                                                      &                                                               &                                      & \begin{tabular}[c]{@{}l@{}}Dena Paona\\ (1923)\end{tabular}      & \begin{tabular}[c]{@{}l@{}}Social, \\ Romantic\end{tabular} & {\color[HTML]{00009B} \textbf{F-A2}} \\ \cline{7-9}
                                                                    &                                                                &                                      &                                                                      &                                                               &                                      & \begin{tabular}[c]{@{}l@{}}Nababidha\\ (1924)\end{tabular}       & \begin{tabular}[c]{@{}l@{}}Social, \\ Romantic\end{tabular} & {\color[HTML]{9A0000} \textbf{M-A2}} \\ \cline{7-9}
                                                                    &                                                                &                                      &                                                                      &                                                               &                                      & \begin{tabular}[c]{@{}l@{}}Pather Dabi\\ (1926)\end{tabular}     & {\color[HTML]{009901} Political}                            & {\color[HTML]{9A0000} \textbf{M-A2}} \\ \cline{7-9}
                                                                    &                                                                &                                      &                                                                      &                                                               &                                      & \begin{tabular}[c]{@{}l@{}}Shesprasna\\ (1931)\end{tabular}      & Social                                                      & {\color[HTML]{00009B} \textbf{F-A2}} \\
                                                                   \hline
\end{tabular}
\caption{Year-wise story, genre and protagonist information for all writers.}
\label{tab-protagonist}
\end{table*}

\section{Detailed Findings}

We discuss some additional findings from our character interaction graph analysis from various perspectives.
\subsection{Age and Gender Distribution}

\begin{table}[h]
\centering
\begin{tabular}{l|ll|lll}
\hline
Author                      & M   & F & A1     & A2     & A3     \\
\hline
BC            & 4.14          & \textbf{5.73}  & \textbf{6.16} & 5.01         & 4.82 \\
RT        & \textbf{4.64} & 3.93          & 4               & \textbf{4.25}  & 3.86 \\
SC & 3.74          & \textbf{5.75} & 4.68          & \textbf{5.30} & 4.22 \\
HM              & \textbf{5.10}  & 4.89         & \textbf{5}      & 4.19           & 4.12 \\
SG         & 4.73          & \textbf{5.58} & 4.16          & \textbf{5.31} & 3.96 \\
\hline
\end{tabular}
\caption{Average Degree Count of Different Age and Gender Group.}
\label{tab-degree}
\end{table}

Average degree count of different age \& gender groups is shown in Table \ref{tab-degree}.
A3 age group also has the least connectivity in all authors (lowest degree count). Urban-centric plot allows male characters in Rabindranath to have more connectivity than female. Also, A2 group has the highest degree count for both Rabindranath and Sarat in previous times because of their more contemporary plots.

\begin{table*}[h]
\centering
\begin{tabular}{lllllll}
\hline
Author                      & M-A1   & F-A1    & M-A2    & F-A2    & M-A3    & F-A3    \\
\hline
Bankim Chandra              & 2.4096 & 14.8148 & 49.3976 & 66.6667 & 48.1928 & 18.5185 \\
Rabindranath Tagore         & 1.3333 & 2.439   & 68.0    & 73.1707 & 30.6667 & 24.3902 \\
Sarat Chandra Chattopadhyay & 7.1429 & 16.1765 & 61.6883 & 58.8235 & 31.1688 & 25.0    \\
Humayun Ahmed               & 1.7857 & 20.0    & 56.25   & 56.3636 & 41.9643 & 23.6364 \\
Sunil Gangopadhyay          & 4.4944 & 8.0     & 79.5506 & 88.0    & 15.9551 & 4.0   \\
\hline
\end{tabular}
\caption{Age and Gender Wise Combined Distribution for All Authors. All Values Are Indicated in Percentage (\%).}
\label{tab-age_gender_combined}
\end{table*}

Table \ref{tab-age_gender_combined} shows the age \& gender-wise combined distribution for all authors. We observe that A3 age group contains more males than females and vice versa for A1 group. Therefore, males in A1 age group are represented as children in most stories, where females in that group can show different roles (lover, widow, married, and others) that assert their higher proportion. Similarly, A3 age group mostly functions as the supporting characters, parents, or moral representation of society that incorporates more male characters due to the patriarch society of contemporary times. Bankim has the almost same ratio of A2 and A3 age group for male characters. Because Bankim's novels mostly incorporate a feudal society background, where kings, landlords, and other influential characters belong to A3 class. All three previous authors have nearly non-existent male characters as A1 age group that implies that children do not receive much attention in the social structure of that period. 
Female characters in A1 group receive attention because of their roles other than as children.
Contemporary authors also show a similar pattern where presence of male children is significantly lower. It demonstrates that male children have always received less attention in Bengali literary fiction.

\begin{table*}[h]
\centering
\begin{tabular}{llll|llllll}
\hline
Author & M-M             & M-F             & F-F    & A1-A1  & A1-A2  & A1-A3  & A2-A2           & A2-A3           & A3-A3  \\
\hline
BC     & 0.3423          & \textbf{0.4878} & 0.1698 & 0.0084 & 0.1204 & 0.0532 & 0.3053          & \textbf{0.4006} & 0.1204 \\
RT     & 0.4328          & \textbf{0.4477} & 0.1193 & 0      & 0.0192 & 0.0115 & \textbf{0.4981} & 0.3793          & 0.092  \\
SC     & 0.4216          & \textbf{0.4612} & 0.1170 & 0.014  & 0.1012 & 0.0471 & \textbf{0.4887} & 0.2914          & 0.0716 \\
HM     & 0.3561          & \textbf{0.4928} & 0.1509 & 0.0088 & 0.0936 & 0.0643 & 0.3216          & \textbf{0.4152} & 0.1053 \\
SG     & \textbf{0.4984} & 0.4264          & 0.0750 & 0      & 0.0511 & 0.015  & \textbf{0.6727} & 0.2282          & 0.033 \\
\hline
\end{tabular}
\caption{Age-wise edge distribution in different authors }
\label{tab-edge_distribution}
\end{table*}

We also observe the interaction between different gender and age group in Table \ref{tab-edge_distribution}. The proportion of male-female link is highest for most authors except Sunil since his plots lack female characters and they have a more cornered role. For previous authors, Rabindranath and Sarat follow a nearly same distribution of gender-wise links. Also, the ratio of A2-A2 (the edge between two middle-aged/young characters) is highest in the age-wise distribution for them, where A2-A3 is the most present edge type for Bankim.  Rabindranath's urban-centric plot and Sarat's rural society-based plot both allocate male and female A2 group as central characters and they do not interact much with the A3 age group as in the feudal social structure of Bankim's novels. Humayun Ahmed's plots also evolve around contemporary social life, family struggle, and therefore also demonstrates A2-A3 type as the highest edge group. Relation among children is very rare for all authors since none of these are children specific fiction.  Also, Bankim and Sarat show a higher A1-A2 ratio than Rabindranath since many of their A1 characters actively participate in the story plot.

\subsection{Influence of Family}

\begin{figure*}[h]
	\centering
	\begin{subfigure}{0.66\columnwidth}
		\centering
		\includegraphics[height=4cm]{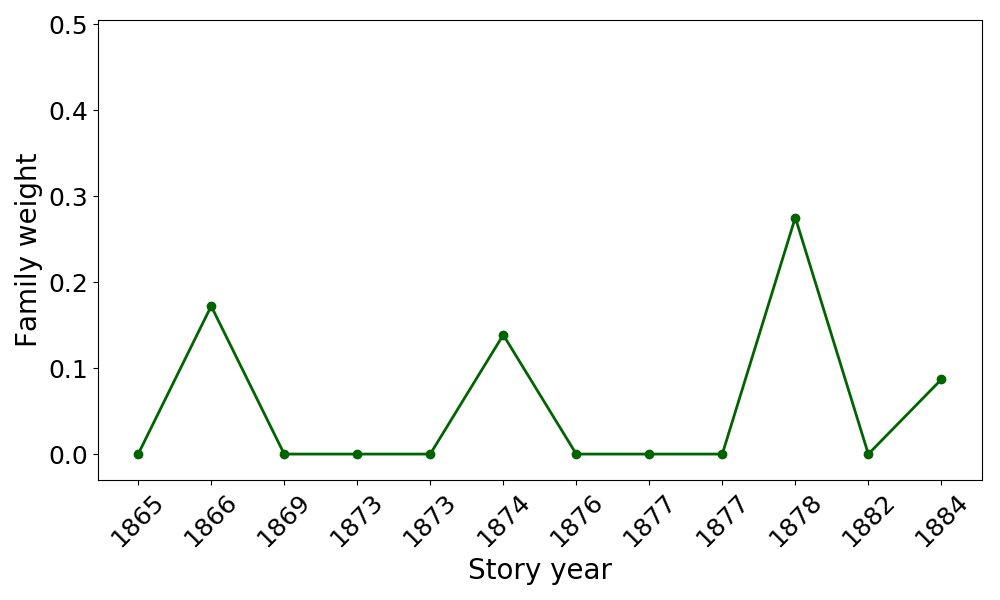}
		\caption{Bankim Chandra}
	\end{subfigure}%
	\begin{subfigure}{0.66\columnwidth}
		\centering
		\includegraphics[height=4cm]{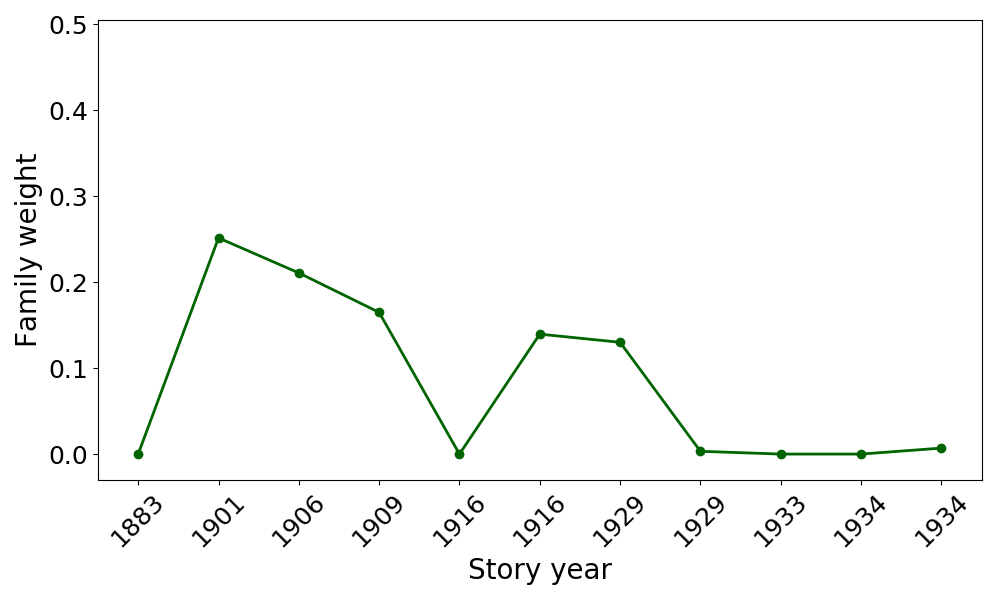}
		\caption{Rabindranath Tagore}
	\end{subfigure}%
		\begin{subfigure}{0.66\columnwidth}
		\centering
		\includegraphics[height=4cm]{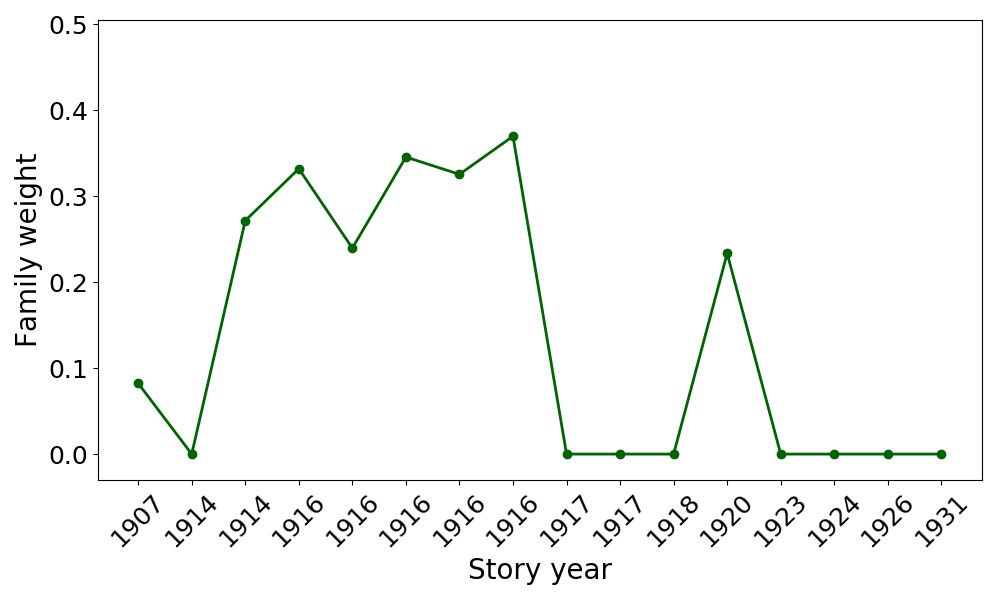}
		\caption{Sarat Chandra Chattopadhyay}
	\end{subfigure}%
	\caption{Weight associated with family members for different authors over time.
	}
	\label{fig-family}
\end{figure*}

Family plays a significant part in the context of the Bengali social structure~\cite{inden2005family}. We intend to examine how, during the era we are studying, the significance of family varies over time in Bengali literature. Therefore, we plot the story-wise average for the aggregated weight of characters with distinct family attributes over time in Figure \ref{fig-family} for all previous authors. In Bankim's writings, family involvement was infrequent since many of his novels follow the background of feudal society. It creates a social network of related individuals who are not family members of the protagonist, revolving around the landlord's character. In the writings of both Rabindranath and Sarat, up to 1916, we observe a greater family weight. After that, in their writings, we show hardly any family presence, especially those novels associated with contemporary social and political problems.

\begin{table*}[h]
\centering
\begin{tabular}{l|l}
\hline
Year      & Event     \\                                    \hline                                                                                                                                                                        
1872      & The Brahma Marriage Act passed lifting ban on widow remarriage                                                                                                                                                                               \\
1882      & Women enter university                                                                                                                                                                                                                       \\
1885      & Bengal Tenancy Act                                                                                                                                                                                                                           \\
1905      & First Partition of Bengal; spearheading the ‘Quit India’ Movement                                                                                                                                                                            \\
1911      & Annulment of Bengal Partition                                                                                                                                                                                                                \\
1912      & Imperial capital shifted from Calcutta to Delhi                                                                                                                                                                                              \\
1914-1918 & Fall in jute prices affects the Bengal economy and  WW-I  effect                                                                                                                                                                                      \\
1917      & Influence of Bolshevism in Bengal intelligentsia                                                                                                                                                                                             \\
1919      & Jallianwala Bagh massacre                                                                                                                                                                                                                    \\
1920      & Non-cooperation movement against colonial rulers led by Gandhi; and civil   disobedience                                                                                    \\
1930      & Salt march movement by Gandhi  \\
\hline
\end{tabular}
\caption{Political and social events in contemporary time of late nineteenth and early twentieth century }
\label{tab-events}
\end{table*}

\section{Case study: Effect of Widow Remarriage Law}
Bankim had widows as his central concern, as he also wrote about widows in a number of his non-fictional
writings and addressed the issue of their remarriage. After the remarriage act, we observe
a significant increase in the female A1 character category in Bankim’s writings. Widows are the central characters in his two novels, Bisabrksa (The Poison Tree, 1872) and Krishnakanter
Will (The Will of Krishnakanta, 1878).

\paragraph{Bisabrksa (1872): }
The major interactions that happen in this story are between the protagonist Nagendra(Index 0, M-A2), one male character Debendra (Index 3, M-A2), four other female characters where three are from A1 age groups, and two are a widow. While the novel's protagonist is Nagendra, the widow character Kundra (Index 1, F-A1) has a higher weight in the story. The inclusion of this widow character creates a triangle of relationship between Nagendra, his wife Surjamukhi  (Index 2, F-A2). Interestingly, the weight of Nagendra-Kundra relationship is greater and more positive than the relationship with his wife.  Also, the other widow character Hira (Index 10, F-A1) receives significant attention and develops interaction with different male characters in the story. The protagonist Nagendra shows 92\% degree connectivity, which conforms to his role as a landlord in contemporary society.

\begin{figure}[h]
	\centering
	\begin{subfigure}{0.45\columnwidth}
		\centering
		\includegraphics[width=\textwidth]{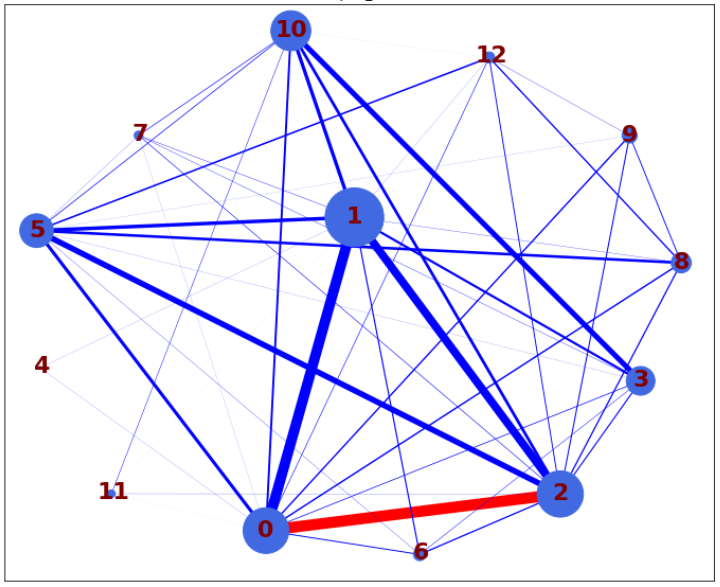}
		\caption{\textit{Bisabrksa (The Poison Tree, 1872)}}
		\label{subfig-brishbikkha}
	\end{subfigure}%
	\begin{subfigure}{0.45\columnwidth}
		\centering
		\includegraphics[width=\textwidth]{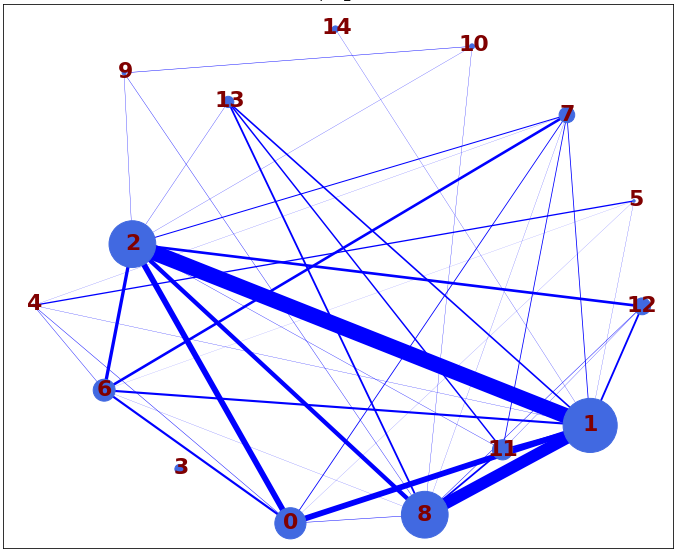}
		\caption{\textit{Krishnakanter Will (The Will of Krishnakanta, 1878)}}
		\label{subfig-krisnokanto}
	\end{subfigure}%
	\caption{Character interaction graph for two novels of Bankim. Protagonist (Index 0) of both stories are widow.
	}
	\label{fig-widow}
\end{figure}

\paragraph{Krishnakanter Will (1878):} 
Widow characters are represented in this social fiction as well. Here the central widow character Rohini (Index 2, F-A2) also serves as the other woman in the triangle with Gobindalal (Index 1, M-A2) and his wife Bhramar (Index 8, F-A1). The weight of the relationship between Gobindalal and Rohini is slightly greater than that of his wife, Bhramar. Gobindalal demonstrates less degree of connectivity because of not being a landlord.

\section{Case Study: Impact of Nationalist Movement}
\begin{figure}[h]
	\centering
	\begin{subfigure}{0.45\columnwidth}
		\centering
		\includegraphics[width=\textwidth]{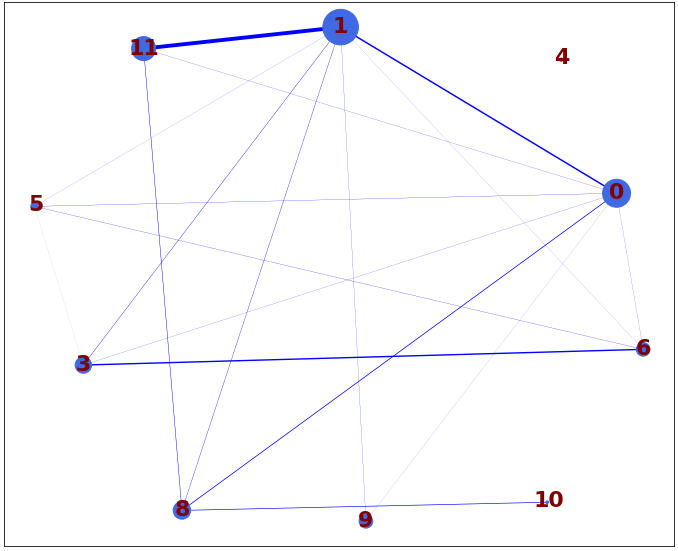}
		\caption{\textit{Ghare Baire (Home and Abroad, 1916)}}
		\label{subfig-ghare_baire}
	\end{subfigure}%
	\begin{subfigure}{0.45\columnwidth}
		\centering
		\includegraphics[width=\textwidth]{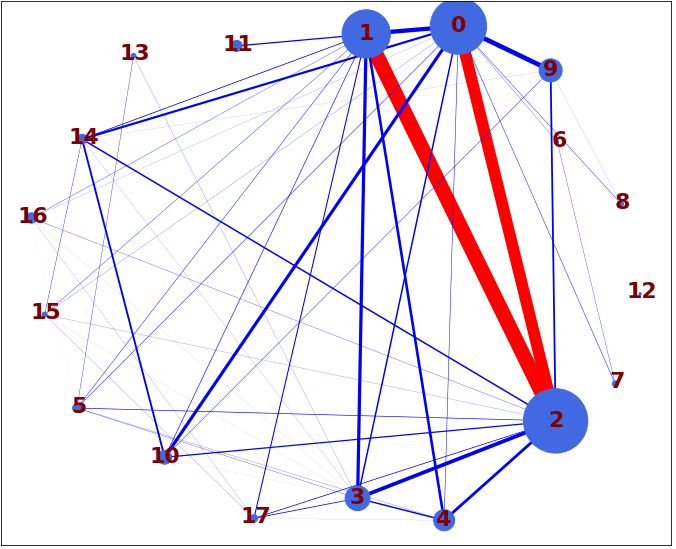}
		\caption{\textit{Pather Dabi (Claim the Path, 1926)}}
		\label{subfig-pather_dabi}
	\end{subfigure}%
	\caption{Character interaction graph for two political novels of Rabindranath and Sarat influenced by nationalist movement. 
	}
	\label{fig-nationalist}
\end{figure}

In the first few decades of the nineteenth century, various nationalist movements influenced Rabindranath and Sarat to portray their perceptions through  fiction. 
To understand whether it is possible to depict contemporary social structures, we briefly review the character interaction graphs of two political novels:  Ghare Baire (Home and Abroad, 1916) by Rabindranth (Figure \ref{subfig-ghare_baire}) and Pather Dabi (Claim the Path, 1926) by Sarat fiction (Figure \ref{subfig-pather_dabi}), which are inspired by non-cooperation movements and other ongoing activities in Indian sub-continent.

\paragraph{Ghare Baire (1916):} 
The plot includes three primary characters, Bimla as the protagonist (Index 0, F-A2), her husband Nikhil (Index 3, M-A2), and a nationalist character Sandeep (Index 1, M-A2). Although Nikhil is described as the hero of the novel by critical analysis~\cite{gupta2016rabindranath_works}, he received less attention in graph. Bimla also does not show the highest node weight, which is achieved by Sandeep. However, Bimla has a higher degree count indicating her connecting with more characters (most of them have some family attributes). Although Bimla was not a part of the nationalist movement, the relationship weight between herself and Sandeep is higher than her relationship with Nikhil. Therefore, this novel represents a small part of contemporary society where a typical female has to maintain a lot of relations with her family members but leans to an iconic figure rather than her husband. Also, the relatively small weight of the female protagonist can be explained by her non-participation in these movements. Later books of Rabindranath portrays higher weight for female protagonists in social/political novels when female participation in national movements was more common~\cite{sarkar1987nationalist}.

\paragraph{Pather Dabi (1926):} 
The plot revolves around a fictional order called "Pather Dabi" and involved primary characters are the protagonist Apurba (Index 0, M-A2), leader of the group Sabyasachi (Index 1, M-A2),  Bharati (Index 2, F-A2), Sumitra (Index 3, F-A2), Tewari (Index 9, M-A2). Except for Sumitra, all major characters are connected to the organization. The negative emotion between Apurba-Bharati and Sabyasachi-Bharati indicates the tension between the group since they are not romantically involved~\cite{chatterjee2009women3}. Although not the leader, protagonist Apurba has more degree count than Sabyasachi. It is also supported by the plot of the story that Apurba later worked as the spy for the police. Several minor characters are a part of the order and form a relationship between them. Therefore, the character interaction graph shows the overall structure of a fictional nationalist organization of contemporary times.

Only one character in Ghare Baire was involved in the nationalist movement, where many characters in Pather Dabi are related to the organization. As a result, Ghare Baire shows less node count as well as many trivial edges with less weight. The node count, graph density, and overall edge weights are higher in Pather Dabi because of its context. 
Therefore, certain historical events have always inspired authors of contemporary times to represent their ideas through fiction. However, we can not confirm that corresponding social structures will always be reflected in this fiction. Rather they depict a very small part of the society, which may include some specific types of characters inspired by real events as we observed in our case studies. How these characters interact in the fictional social network, however, does not always depend on the precise social structure. It depends, instead, on the perception of these events by the authors and their imagination.

\end{document}